\documentclass{article}

\usepackage{arxiv}
\usepackage[utf8]{inputenc} 
\usepackage[T1]{fontenc}    
\usepackage{hyperref}       
\usepackage{url}            
\usepackage{booktabs}       

\usepackage{nicefrac}       
\usepackage{microtype}      
\usepackage{lipsum}		
\usepackage{graphicx}
\usepackage{subfigure}
\usepackage{doi}
\usepackage{xcolor}
\usepackage{amsmath,amssymb,amsfonts}
\usepackage[ruled,vlined,linesnumbered]{algorithm2e}
\usepackage{mathrsfs}
\usepackage[square,numbers,sort&compress]{natbib}
\DeclareMathOperator{\ODEGenerator}{\textit{ODEGenerator}}
\DeclareMathOperator{\ODESOLVER}{\textit{ODESolver}}

 \DeclareMathOperator{\ODEECGGenerator}{\textit{ODEECGGenerator}}

\title{ECG synthesise with Neural ODE and GAN models}

\author{ 
	\href{https://orcid.org/0000-0001-9051-1370}{\includegraphics[scale=0.06]{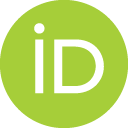}\hspace{1mm}Mansura Habiba} \\
	Dept of Computer Science\\
	Maynooth University\\
	Maynooth, Ireland \\
	\AND
	\href{https://orcid.org/0000-0002-6486-5746}{\includegraphics[scale=0.06]{orcid.png}\hspace{1mm}Eoin Brophy} \\
	School of Computing  \& INFANT Research Centre\\
		Dublin City University\\
	Dublin, Ireland \\
	\AND
	\href{https://orcid.org/0000-0003-0521-4553}{\includegraphics[scale=0.06]{orcid.png}\hspace{1mm}Barak A. Pearlmutter} \\
	Department of Computer Science \& Hamilton Institute\\
	Maynooth University\\
	Maynooth, Ireland \\
	\AND
	\href{https://orcid.org/0000-0002-6173-6607}{\includegraphics[scale=0.06]{orcid.png}\hspace{1mm}Tomas Ward} \\
	School of Computing  \& INFANT Research Centre\\
		Dublin City University\\
	Dublin, Ireland \\
}

\renewcommand{\shorttitle}{\textit{arXiv} Template}

\hypersetup{
pdftitle={Neural ODE based models for Continuous  Medical Time Series Synthesis},
pdfsubject={q-bio.NC, q-bio.QM},
pdfkeywords={Neural ODE, Generative Adversarial Network, ECG synthesis, Time series generation},
}

\begin{document}
\maketitle

\begin{abstract}
 Continuous medical time series data such as ECG is one of the most complex time series due to its dynamic and high dimensional characteristics. In addition, due to its sensitive nature, privacy concerns and legal restrictions, it is often even complex to use actual data for different medical research. As a result, generating continuous medical time series is a very critical research area. Several research works already showed that the ability of generative adversarial networks (GANs) in the case of continuous medical time series generation is promising. Most medical data generation works, such as ECG synthesis, are mainly driven by the GAN model and its variation. On the other hand, Some recent work on Neural Ordinary Differential Equation (Neural ODE) demonstrates its strength against informative missingness, high dimension as well as dynamic nature of continuous time series. Instead of considering continuous-time series as a discrete-time sequence, Neural ODE can train continuous time series in real-time continuously. In this work, we used Neural ODE based model to generate synthetic sine waves and synthetic ECG. We introduced a new technique to design the generative adversarial network with Neural ODE based Generator and Discriminator. We developed three new models to synthesise continuous medical data. Different evaluation metrics are then used to quantitatively assess the quality of generated synthetic data for real-world applications and data analysis.  Another goal of this work is to combine the strength of GAN and Neural ODE to generate synthetic continuous medical time series data such as ECG. We also evaluated both the GAN model and the Neural ODE model to understand the comparative efficiency of models from the GAN and Neural ODE family in medical data synthesis.          
\end{abstract}


\section{Introduction}
Neural Ordinary Differential Equations (NODE) \cite{chen2018neural} introduces a new family of neural network. NODE can generate continuous-time series. On the other hand, Generative Adversarial Network (GAN) models demonstrate impressive performance in synthetic image generation. However, the performance of GAN in terms of synthetic time series generation \citep{ledig2017photo,hartmann2018eeg, li2019mad} is comparatively weak. ECG signal is a complex data structure. ECG signal often exhibits dynamic characteristics, such as irregularity with informative missingness \cite{grud}, high dimension, multi-variate, large in length, high frequency or data sampling rate and dynamic pattern in the dataset. Recent Neural Ordinary Differential Equations based models \citep{chen2018neural, habiba2020neuralode, kidger2020neural,grathwohl2018ffjord, rubanova2019latent} demonstrate improved performance to overcome different challenges in continuous time series. 

Continuous medical time series is susceptible and require additional security. Using real data in medical research is often challenging. We have introduced  NODE and Generative Adversarial Network (GAN) models to generate continuous medical time series data, e.g. electrocardiogram (ECG). These generated data can be used instead of actual ECG without compromising the quality of analysis.

The main objectives of this work are as follows:

\begin{itemize}
	\item Develop NODE based models for continuous medical time series generation
	\item Improve the performance of Generative  Adversarial Neural Network to generate continuous medical time series
	\item Evaluate the performance of proposed models.
\end{itemize}

\section{Background}
GAN models are top-rated for signal synthesis. Several recent works show excellent outcome for signal synthesis problem. Most GAN model for signal synthesis problems is often Recurrent neural networks. In this work, we focus on developing a NODE model for signal synthesis problem. We also introduced a GAN model based on NODE to provide a better result for the signal synthesis problem.  
NODE \cite{che2018recurrent, habiba2020neural},  can be presented as a continuous function instead of several layers, and the hidden state can be parametrised using an Ordinary Differential Equation (ODE) solver.  ODE solvers help to compute the changes in hidden dynamics as shown in \ref{eq:dh_t}. 
\begin{equation}
	\label{eq:dh_t}
	\frac{dh(t)}{dt} =\ODESOLVER(f, h_{t},t, \delta_{t})
\end{equation} 

Here $f$ is an initial problem ODE solver. Using the initial condition of $h(0)$, the ODE solver compute the output $h(t)$ at time $t$. This approach provides faster training time than residual network, constant memory cost instead of linearly increasing memory cost and designing any neural network model is much simpler. 

Research shows that continuous time series or signal modelling can achieve a better result with recurrent neural network \cite{neil2016phased}. If the ODE solver in NODE can leverage the architecture of Recurrent neural network (RNN), it can learn continuous time series in real-time with higher precision\cite{habiba2020neural, habiba2020neuralode}. Eq \ref{eq:neural-odernn} shows that generic NODE based RNN cell \cite{habiba2020neural} is used by ODE solver to compute the output $y_t$ at time $t$ from the initial input $y_0$ at time $t_0$.
\begin{equation}
	y_{t} = \ODESOLVER(\textit{ODERNNCell},y_0, t)
	\label{eq:neural-odernn}
\end{equation}

Several recent works  \cite{habiba2020neural, habiba2020neuralode, che2018recurrent, rubanova2019latent, linial2020generative}shows that NODE models show significantly better performance for continuous-time series (i.e. sine wave, ECG, etc.) generation. In this work, we introduced three models for continuous medical time series synthesis.

\begin{figure*} [http]
  \subfigure[MIT-BIH Normal Sinus]{\includegraphics[width=0.5\textwidth]{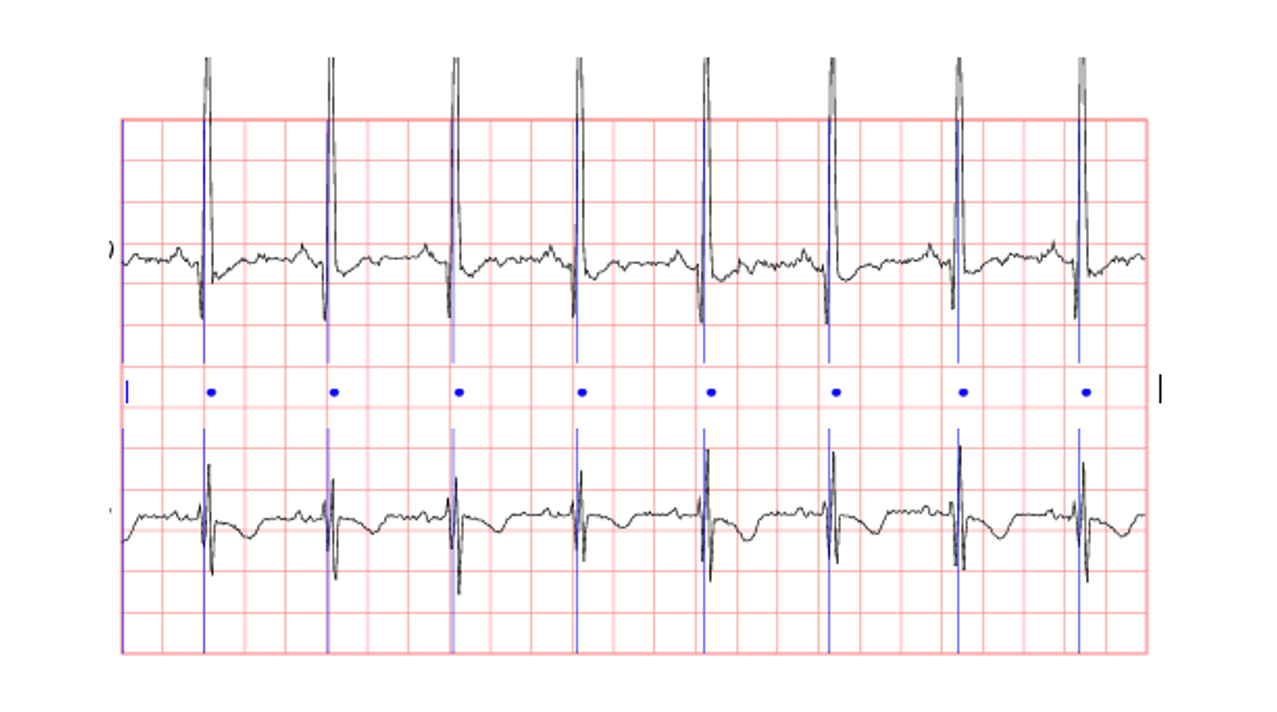}}\label{fig:mit-normal}
\hfill
\subfigure[MIT-BIH Arrhythmia]{\includegraphics[width=0.5\textwidth]{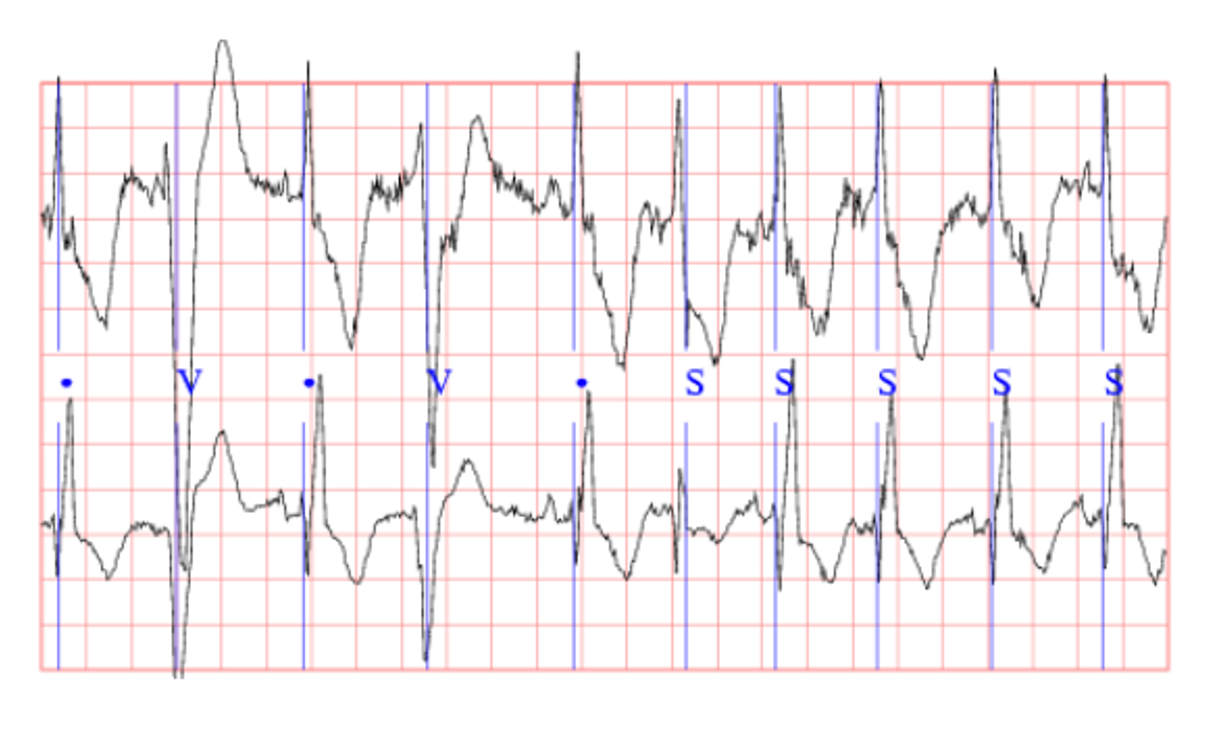}}\label{fig:arr-mit}
\vfill
\caption{Loss analysis for training GAN model}
\label{fig:mit}
\end{figure*}

My research aims to use deep learning to generate synthetic electrocardiogram (ECG) data representative of real ECG. My goal is to generate high-quality data that can reduce the limitation in medical research for lack of data. Fig \ref{fig:mit} shows typical ECG signal. Fig. \ref{fig:mit}~(a) represents Normal Sinus data and Fig. \ref{fig:mit}~(b) represents irregular ECG for Arrhythmia. Both ECG signals are continuous in time with a higher data sampling rate.  

Among different deep learning models, GANs are leading the research for ECG data generation. \citep{brophy2020synthesis,delaney2019synthesis, golany2020simgans, golany2019pgans} shows how GANs generate ECG data. Among these works, \citep{golany2020simgans, golany2019pgans} shows domain-specific knowledge such as personalisation or external simulation can improve the performance of GANs model. But training personalised data with ECG data is time-consuming and needs multiple environments for training. On the other hand, \citep{brophy2020synthesis,delaney2019synthesis} adopt the hybrid model concept to design the architecture for GAN. GANs described in \citep{brophy2020synthesis,delaney2019synthesis} use Recurrent Neural Networks (RNN) as Generator and a combination of sequential RNN as well as convolutional neural networks as the Discriminator. But there are still limitations, as GANs are unstable in training and does not have proper suitable evaluation measures. The dynamical equations of motion for ECG signal can be described as ordinary differential equations \cite{mcsharry2003dynamical}. Therefore, ODE can model ECG signal as a function of continuous-time. Ordinary Differential Equations also help to model the dynamic characteristics of the heart rate , for example the mean and standard deviation of the ECG signal and spectral properties such as the LF/HF ratio. Therefore, NODE can provide better performance than GAN.  In this work, we focus on using NODE to reduce the limitation for GAN and other deep learning models for ECG data generation.

\section{Model Design}
In this work, we focused on exploring the strength of NODE for ECG synthesis. Firstly, we designed a NODE based model to produce synthetic continuous-time data, which is described in section~\ref{sec:modela}. Secondly, we tried to enrich the architecture of the traditional GAN model using NODE. For that, we designed two separate GAN models, where we tried to design the generator and discriminator using NODE. 

\subsection{Neural ODERNN Model (ODEECGGenerator) Design for ECG Synthesis}
\label{sec:modela}
\begin{figure}[htb]
 \centering
 	\includegraphics[width=0.8\textwidth]{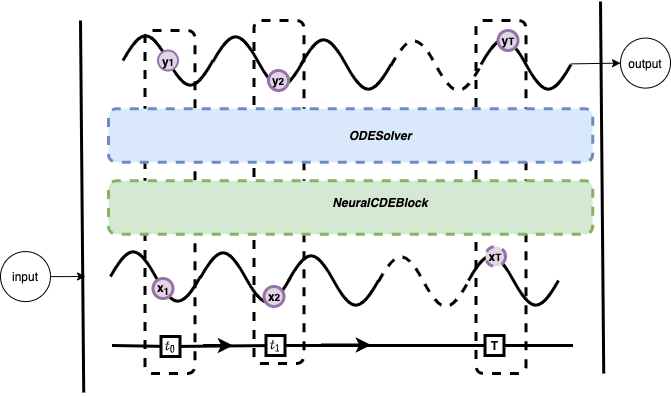} 
 	\label{fig:odeecgg}
 	\caption{The architecture for proposed $\ODEECGGenerator$  model to generate ECG data} 
 \end{figure}

The first model, we designed is an ODE-RNN \cite{habiba2020neural} model, called $\ODEECGGenerator$ . $\ODEECGGenerator$ is simply an Neural ODE-RNN \cite{habiba2020neural} NODE model.  Fig. \ref{fig:odeecgg} shows the architecture for the $\ODEECGGenerator$  model to generate ECG. The building block for this model is a NODE block that leverage Ordinary Differential Equation Solver $\ODESOLVER$ to train an ODEGRU, or ODELSTM model \cite{habiba2020neural}.
 
 $\ODEECGGenerator$  model learns the dynamics of ECG signal in the form of Ordinary Differential Equation (ODE). The proposed Generator in this GAN model, called $\ODEECGGenerator$ , receives ECG signal as a continuous function of time (t). Therefore, input for $\ODEECGGenerator$  is a value at a specific time step $x_{t}$ and produces the derivative of the system at that time step. $\ODEECGGenerator$  takes the initial value $y_0$ for ECG signal at time $t_0$ as shown in Eq. \ref{eq:ch3_odeecg}. $\ODEECGGenerator$  can be described as an ODEn as shown in Eq. \ref{eq:ch5_g11}. The \textit{NeuralODEBlock} in Fig.~\ref{fig:odeecgg} converts the incoming signal and the hidden state of the system as ODE which is passed to the $\ODESOLVER$. $\ODESOLVER$ uses an \textit{ODERNNCell} as shown Eq.~\eqref{eq:neural-odernn}, which is an ODE-RNN  (either \textit{ODEGRU} or \textit{ODELSTM}) model. To solve the ODE within time boundary $[t_{0}, t]$. \textit{ODERNNCell} is optimized using the parameter $\gamma$. The incoming signal contains additional noise (z) to preserve data privacy and keep the model efficient against Adversarial attack. which is also passed as parameter for Eq.~\eqref{eq:neural-odernn}. Therefore,  $\theta = tuple(z, \gamma)$..

\begin{subequations}
\begin{align}
\label{eq:ch3_odeecg}
y_{t}  &= \textit{ODEECGGenerator}(x_t, t)
\\
\label{eq:ch5_g11}
\frac{dy}{dt} &= \textit{NeuralODEBlock}(\ODESOLVER,y_t, t,\theta)
\end{align}
\label{eq:modela}
\end{subequations}

This model is not a GAN model, and rather it is simply an Ordinary Differential Equation based RNN model. An $\ODESOLVER$ solves the Ordinary Differential Equation shown in Eq.~\eqref{eq:ch5_g11} within time $[t_{0}, t]$

\subsection{ODEGAN Model with NODE based Generator and Discriminator}
\label{sec:modelb}
\begin{figure}
    \centering
    \includegraphics[width=0.5\textwidth]{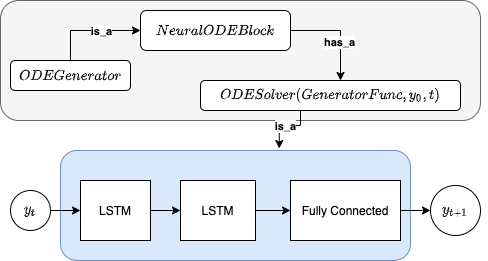}
    \caption{Block Diagram of Generator Architecture}
    \label{fig:model2_gen}
\end{figure}

ODEGAN Model has a NODE generator, called $\ODEGenerator$. The discriminator model is similar to the discriminator model used in \cite{delaney2019synthesis}. Eq.~\eqref{eq:ch5_g12} shows that the \textit{NeuralODEBlock} of $\ODEGenerator$ uses a \textit{GeneratorFunc} as the neural network (f) for $\ODESOLVER$.  \textit{GeneratorFunc} is an recurrent neural network, which can be optimized using parameter $\theta$. Fig.~\ref{fig:model2_gen} shows the architecture for the $\ODEGenerator$ generator. NeuralODEBlock computes the derivative of the input w.r.t. time. Later $\ODESOLVER$ produces a solution for the Ordinary differential equation as shown in Eq.~\eqref{eq:ch5_g12} for time T $[t0, ..., t]$. 
\begin{subequations}
\label{eq:ch5_odegenerator}
\begin{align}
\label{eq:ch5_g1}
y_{t}  &= \ODEGenerator(y_0)
\\
\label{eq:ch5_g12}
\frac{dy}{dt} &= NeuralODEBlock(\textit{GeneratorFunc}, y_0, t, \theta, \textit{noise})
\end{align}
\end{subequations}

\begin{figure}
    \centering
    \includegraphics[width=0.6\textwidth]{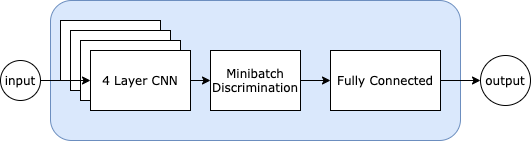}
    \caption{Block Diagram of Generator Architecture}
    \label{fig:model2_des}
\end{figure}

The Discriminator for this GAN model is a four-layer 2-dimensional convolutional neural network and a Minibatch discrimination layer. The discriminator model is similar to the discriminator model used in \cite{delaney2019synthesis}. The discriminator model also has noise added to the gradient of the optimiser to preserve privacy. Fig~\ref{fig:model2_des} shows the architecture for the discriminator model for this model.

Table \ref{tab:desc1} describes the parameter used in the Discriminator model.

\begin{figure}
    \centering
    \includegraphics[width=0.8\textwidth]{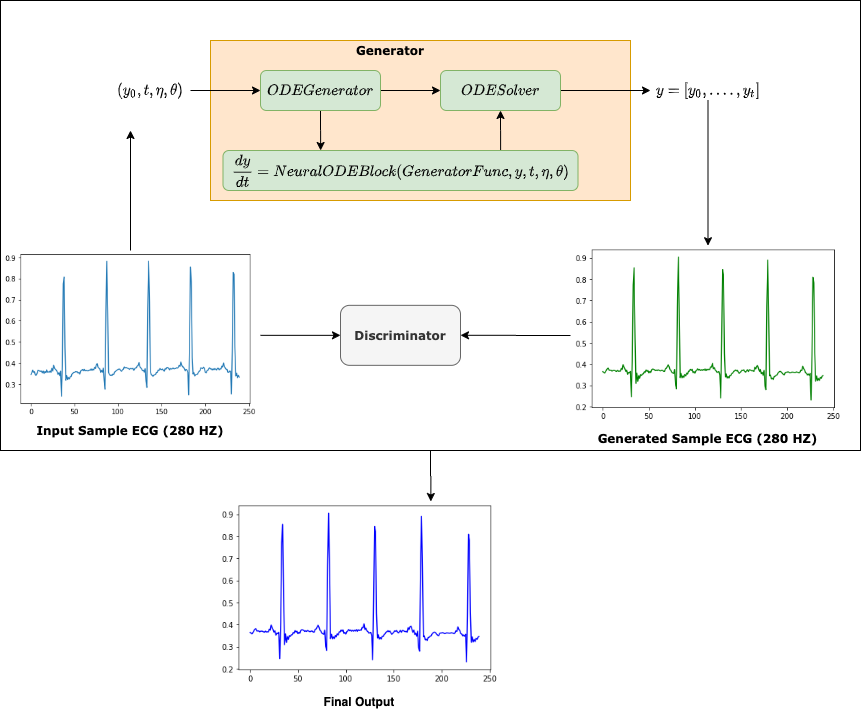}
    \caption{pipeline for GAN model with ODE generator}
    \label{fig:modelb_pipeline}
\end{figure}

\begin{table}[!htb]
	\caption{Parameter used in proposed ODE based Discriminator}
	\centering
	\begin{tabular}{ll}
		\toprule
		seq\_length & Length of the Input Sequence      \\
		batch\_size     & Size of each batch    \\
		minibatch\_normal\_init     & Cell body         \\
		
		num\_cv & Number of convolution Layer. Here it is 2.      \\
		cv1\_out     & Output shape of the first convolution Layer   \\
		cv1\_s     &  Stride shape of the first convolution Layer         \\
		p1\_k     &  Padding length of the first convolution Layer         \\
		cv1\_k     &  Kernel size of the first convolution Layer         \\
		cv2\_out     & Output shape of the second convolution Layer   \\
		cv2\_s     &  Stride shape of second convolution Layer         \\
		p2\_k     &  Padding length of the second convolution Layer         \\
		cv2\_k     &  Kernel size of the second convolution Layer         \\
	    ODEDBlock & The ODE block for the ODE based Discriminator \\
	    ode\_discriminator & the proposed Discriminator \\
		\bottomrule
	\end{tabular}
	\label{tab:desc1}
\end{table}
Fig.~\ref{fig:modelb_pipeline} shows the pipeline for this model. Sample ECG signal $y_{t}$ is passed to the $\ODEGenerator$ model. The initial value for the signal is $y_{0}$. In order to preserve privacy, additional noise $\eta$ is added along with parameter ($\gamma$) of the \textit{GeneratorFunc}, therefore, $\theta = tuple(z, \gamma)$. NeuralODEBlock converts the signal to ODE, and then $\ODESOLVER$ solves the ODE, yielding a generated noisy ECG signal for time T $[t0, ..., t]$. The Discriminator then learn to distinguish between the real and the generated signal. 

During the training phase, $\ODEGenerator$ generates continuous time series similar to a real ECG signal. A log probability (\texorpdfstring{$\mathcal{L}$}) matrix evaluates the $\ODEGenerator$ model by computing $\mathcal{L}$ of negative identification of the generated signal by the  Discriminator. The optimiser optimises the parameter of the  $\ODEGenerator$ model with a target to minimise $\mathcal{L}$. On the other hand, a cross-entropy loss function evaluates the Discriminator by computing $\mathcal{L}$ of correct classification of the real and generated signal. During the training, the optimiser designated for the Discriminator reduces $\mathcal{L}$ to its minimum value.

\subsection{GAN Model with NODE based Generator and Discriminator}
\label{sec:modelc}
For this ODE-GAN-2 model, we designed both Generator and Discriminator using NODE models. The generator model described in section~\ref{sec:modela} is the Generator for this generative adversarial neural network. Eq.~\eqref{eq:modela} shows that the NeuralODEBlock uses an ODERNNCell, which can be either LSTM or GRU to generate a continuous time series $y_{t}$ by using an $\ODESOLVER$ on the ODE defined in Eq.~\eqref{eq:ch5_g11}.

For this model, we have used two different kinds of Discriminator as follows.

\begin{itemize}
    \item Convolution layer with NODE layer
    \item NeuralCDE network  \cite{kidger2020neural}
\end{itemize}

\subsubsection{Discriminator with Convolution layer with NODE layer}

\begin{figure}
    \centering
    \includegraphics[width=0.75\textwidth]{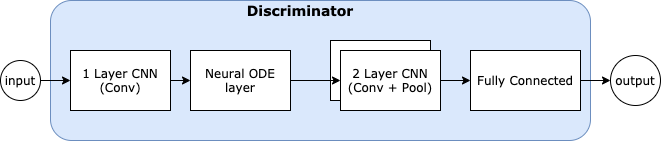}
    \caption{Block Diagram for the Discriminator}
    \label{fig:ode_des_model3}
\end{figure}

This Discriminator has one convolutional layer followed by a NODE layer and then two more convolutional and max-pooling pair layers.  Fig~\ref{fig:ode_des_model3} shows the block diagram of the Discriminator.

Table \ref{tab:desc1} describes the parameter used in proposed ODE based Discriminator. $DiscriminatorFunc$ is the neural network to learn the difference between real and generated ECG signal. $DiscriminatorFunc$ also used by the $\ODESOLVER$ in the Discriminator. This Discriminator distinguishes between the derivative of ECG signal w.r.t. time of the real and generated signal.

\subsubsection{NeuralCDE network as Discriminator}
\begin{figure}[!htb]
    \centering
    \includegraphics[width=0.6\textwidth]{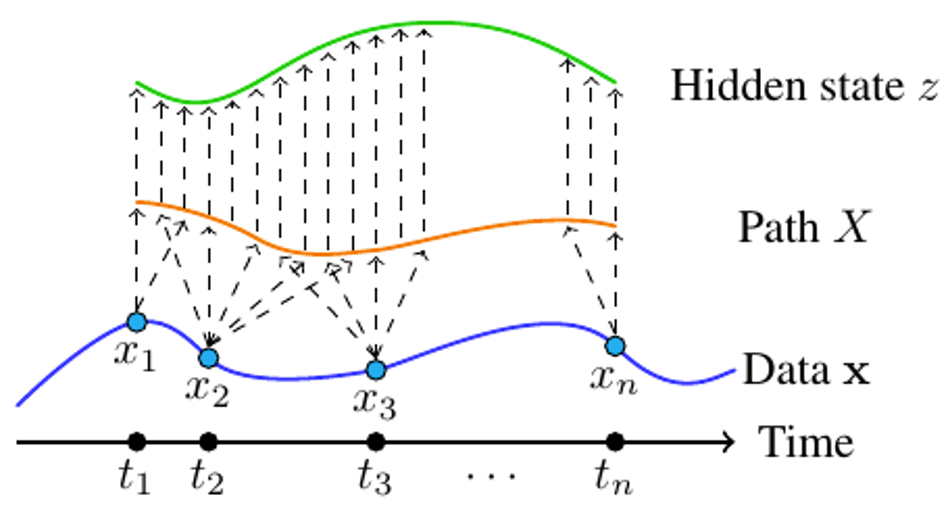}
    \caption{NeuralCDE model \cite{kidger2020neural}}
    \label{fig:neural_cde}
\end{figure}

The Discriminator for this model is a NeuralCDE network  \cite{kidger2020neural}. NeuralCDE network as shown in Fig~\ref{fig:neural_cde}, convert the data to a conditional continuous path $X$ using interpolation and this path $X$ is passed through $\ODESOLVER$ to solve the ODE derived from path $X$ in order to lean the hidden dynamics of the ODE system. For my third model, we leverage the concept of the NeuralCDE network to learn the difference between a real ECG signal and the generated ECG signal. The Discriminator model described in Algorithm \ref{alg:desc2} uses a Neural controlled differential equations \cite{kidger2020neural} as the $\ODESOLVER$ function. Fig. ~\ref{fig:neural_cde} shows that the input time series for this Discriminator has two-channel, e.g. for real ECG signal and generated ECG signal.

\begin{algorithm}[htb]
	\label{alg:desc2}
	discriminatorFunc = NeuralCDE(input\_channels=seq\_length, hidden\_channels=hidden\_dim, output\_channels=out\_dim)
	ode\_discriminator = nn.Sequential(discriminatorFunc)
	\caption{The Neural CDE based Discriminator}
\end{algorithm} 

Table \ref{tab:desc3} describes the parameter used in proposed NeuralCDE based Discriminator.

\begin{table}[htb]
	\caption{The Parameters of NeuralCDE based Discriminator}
	\centering
	\begin{tabular}{ll}
		\toprule		
		input\_channels     &  Length of the Input Sequence     \\
		
		hidden\_dim & Hidden dimension of the  NeuralCDE network   \\
		output\_channels     & Output dimension of the  NeuralCDE network    \\	
		\bottomrule
	\end{tabular}
	\label{tab:desc3}
\end{table}

\begin{figure}
    \centering
    \includegraphics[width=0.75\textwidth]{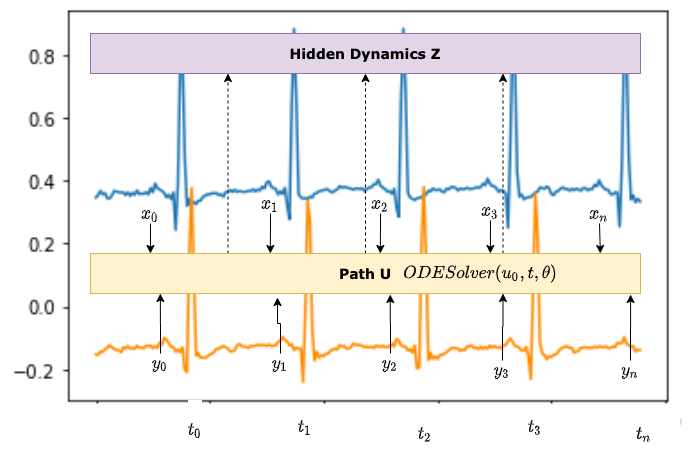}
    \caption{NeuralCDE based Discriminator}
    \label{fig:neural_cde_desc}
\end{figure}

Fig.~\ref{fig:neural_cde_desc} shows that the discriminator takes real ECG signal $x_{t}$ and generated  ECG signal $y_{t}$ as input. Both signals $x_{t}$ and $y_{t}$ are controlled ordinary differential equations.  interpolation between $x_{t}$ and $y_{t}$ generates an intermidieate continuous path $U$. NeuralCDE based Discriminator learns the hidden dynamics ($Z$) of $U$ to distinguish real signal and generated signal correctly. 

\section{Performance Evaluation}
\label{sec:experiments}
The quality and performance of the proposed model are assessed using an experiment with two different tasks, i.e. (i) Generate Normal Sinus ECG and (ii) Generate Arrhythmia ECG. The proposed model is evaluated against state art Neural Network used for ECG Synthesis as described in \cite{delaney2019synthesis} and NeuralCDE.

\subsection{Dataset}
This experiment uses two different kinds of multi-channel publicly available ECG datasets, e.g. (i)   MIT-BIH Arrhythmia dataset on PhysioNet \cite{goldberger2000physiobank} and  (ii) MIT-BIH Normal Sinus Rhythm Database \cite{goldberger2000physiobank}. Both of the Databases show unique characteristics. For example,  MIT-BIH Normal Sinus Rhythm Database consists of clean ECG recordings. As shown in Fig.~\ref{fig:mit-normal}, this Database shows minimal noise in the ECG continuous series. On the other hand, the ECG recordings for MIT-BIH Arrhythmia on PhysioNet were created by adding calibrated noise to clean the MIT-BIH Normal Sinus Rhythm Database. Therefore, the MIT-BIH Arrhythmia dataset contains a significant amount of noise, as shown in Fig.~\ref{fig:arr-mit}.

\begin{figure*}[http]
\hfill
\subfigure[ MIT-BIH Normal Sinus Rhythm Database First Channel]{\includegraphics[width=0.45\textwidth]{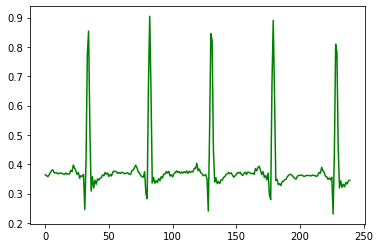}}\label{norm-first}
\hfill
\subfigure[ MIT-BIH Normal Sinus Rhythm Database Second Channel]{\includegraphics[width=0.45\textwidth]{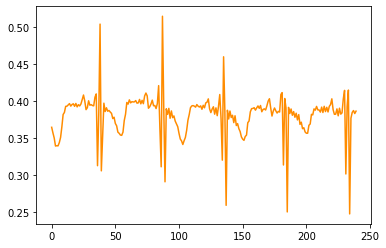}}\label{norm-second}
\vfill
\subfigure[ MIT-BIH Arrhythmia dataset on PhysioNet First Channel]{\includegraphics[width=0.5\textwidth]{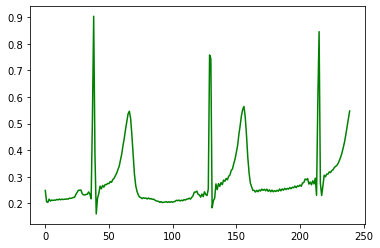}}\label{arrhythmia-first}
\hfill
\subfigure[ MIT-BIH Arrhythmia dataset on PhysioNet Second Channel]{\includegraphics[width=0.5\textwidth]{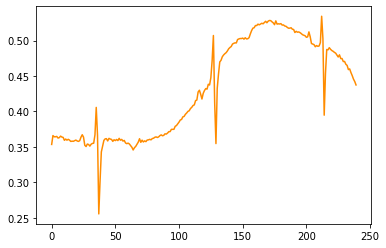}}\label{arrhythmia-second}
\caption{Configuration for neural network used in Human activity experiment}
\label{fig:dataset}
\end{figure*}

On the other hand, the MIT-BIH Arrhythmia dataset on the PhysioNet database has a significant amount of noise. The ECG recordings for MIT-BIH Arrhythmia on PhysioNet were Database was created by adding calibrated amounts of noise to clean ECG recordings from the MIT-BIH Normal Sinus Rhythm Database.

\subsection{\texorpdfstring{$\ODEECGGenerator$} model Training}
\begin{figure}[htb]
    \centering
    \includegraphics[width=0.5\textwidth]{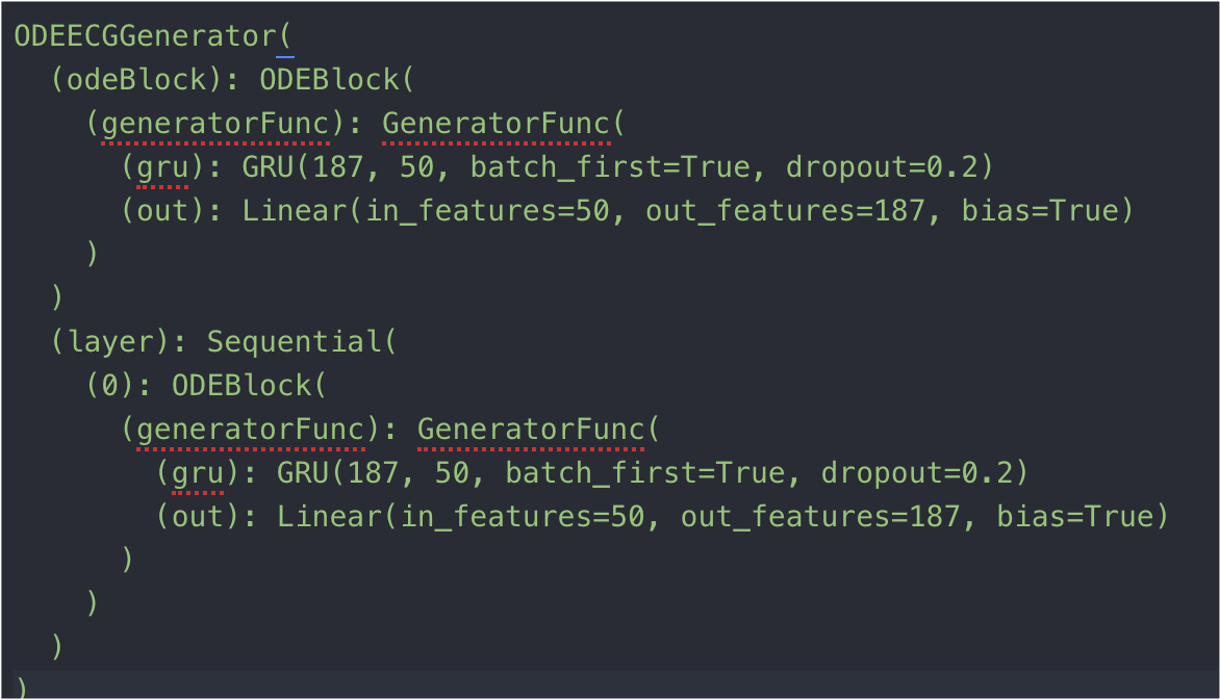}
    \caption{Proposed $\ODEECGGenerator$ model for ECG Synthesis}
    \label{fig:model_ode}
\end{figure}

Fig. \ref{fig:model_ode} shows the architecture for proposed $\ODEECGGenerator$ model used in this experiment. 

Table \ref{tab:modela} describes the settings used for performance evaluation. The dynamic characteristics of proposed models enable us to train them quickly and with fewer parameters. For robust training, the dataset was split at random into  80\% for training and 20\% for the test.

\begin{table}[htb]
	\caption{The Parameters of Experiment for $\ODEECGGenerator$ and GAN models training}
	\centering
	\begin{tabular}{lll}
		\toprule
		Parameters & $\ODEECGGenerator$ & GAN \\
		Batch Size     &  50  & 64  \\
		dataseize & 100 & 1000\\
		Sequence Length & 240  & 240  \\
		Hidden Dimension     & 50  & 50  \\	
		Learning rate    & 0.0001  & 0.00005  \\
		Number of epoch    & 100  & 30 * 1000 \\
		\bottomrule
	\end{tabular}
	\label{tab:modela}
\end{table}

\subsection{GAN model for ECG Synthesis}
\begin{figure}[htb]
    \centering
    \includegraphics{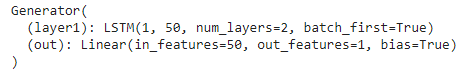}
    \caption{Generator model architecture}
    \label{fig:gan_model}
\end{figure}

For comparative analysis, the proposed model is evaluated against the GAN model for ECG synthesis described in \cite{delaney2019synthesis}. Fig \ref{fig:gan_model} shows the block diagram for Generator and Discriminator used in \cite{delaney2019synthesis}.

\begin{figure*}[http]
\hfill
\subfigure[Block Diagram of Generator Architecture]{\includegraphics[width=0.45\textwidth]{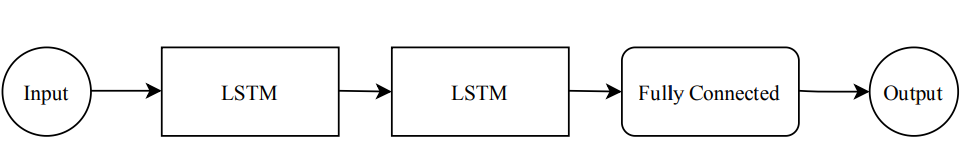}}
\hfill
\subfigure[Block Diagram of Discriminator Architecture]{\includegraphics[width=0.45\textwidth]{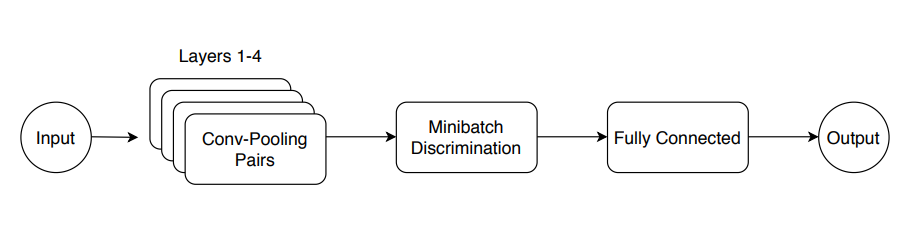}}
\caption{Block Daigram for Generator and Discriminator for GAN model for ECG synthesis described in \cite{delaney2019synthesis}}
\label{fig:gan}
\end{figure*}

The Generator for GAN model \cite{delaney2019synthesis} as shown in Fig. \ref{fig:gan_model} has two consecutive RNN layers (LSTM), followed by a fully connected layer.

\begin{figure}[htb]
    \centering
    \includegraphics[width=0.85\textwidth]{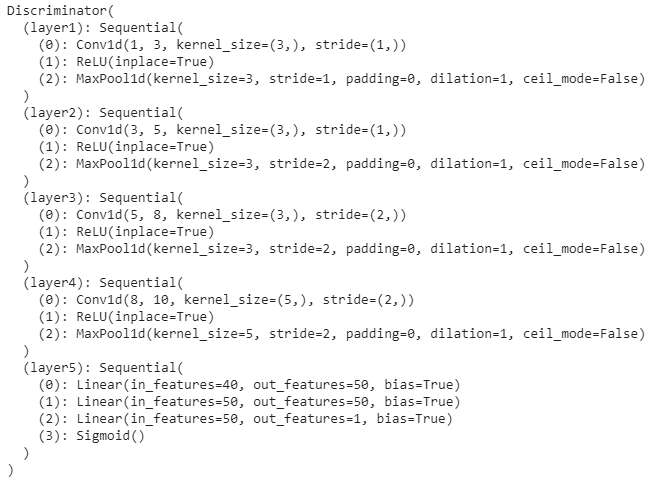}
    \caption{Discriminator model architecture}
    \label{fig:desc_model}
\end{figure}

The Discriminator proposed in \cite{delaney2019synthesis} has a comparatively complex architecture as shown in Fig. \ref{fig:desc_model} and Fig. \ref{fig:gan}.

Table \ref{tab:desc1} describes the parameter used in GAN implementation of \cite{delaney2019synthesis}. GAN model training requires a comparatively higher number of parameters than the proposed  $\ODEECGGenerator$ model. It also needs to be trained for a longer time for each epoch. Out of 30 epochs runs through 1000 iteration. A comparatively longer series of 1000 data size is required for the GAN model. The training requires three to five hours for 30 epoch. 

\begin{figure*}[htb]

\subfigure[Original MIT-BIH Normal Sinus Rhythm  ]{\includegraphics[width=0.45\textwidth]{./images/normal_sinus_first_channel}}
\hfill
\subfigure[ $\ODEECGGenerator$ generated MIT-BIH Normal Sinus Rhythm  ]{\includegraphics[width=0.45\textwidth]{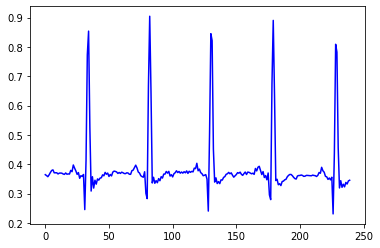}}
\vfill
\subfigure[Original MIT-BIH Arrhythmia First Channel]{\includegraphics[width=0.45\textwidth]{./images/arrhythmia_first_channel}}
\hfill
\subfigure[ $\ODEECGGenerator$ generated MIT-BIH Arrhythmia First Channel]{\includegraphics[width=0.45\textwidth]{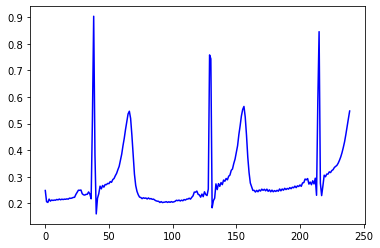}}
\vfill
\subfigure[GAN generated MIT-BIH Arrhythmia First Channel ]{\includegraphics[width=0.45\textwidth]{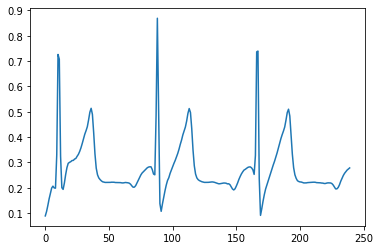}}
\hfill
\subfigure[ $\ODEECGGenerator$ generated MIT-BIH Arrhythmia First Channel]{\includegraphics[width=0.45\textwidth]{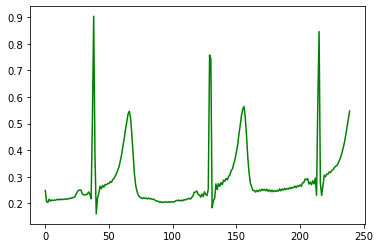}}
\vfill
\caption{Comparative result analysis for proposed model}
\label{fig:generated-signal}
\end{figure*}

\begin{figure*} [http]
  \subfigure[ $\ODEECGGenerator$ model training loss for MIT-BIH Normal Sinus Rhythm Database  First Channel]{\includegraphics[width=0.45\textwidth]{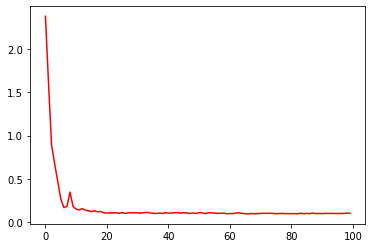}} \label{fig:loss_normal_f}
\hfill
\subfigure[ $\ODEECGGenerator$ model training loss for MIT-BIH Normal Sinus Rhythm Database  Second Channel]{\includegraphics[width=0.45\textwidth]{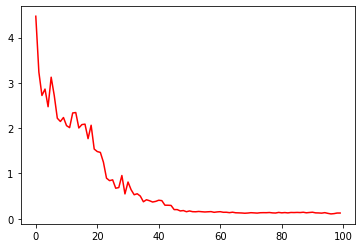}}\label{fig:loss_normal_s}
\vfill
\subfigure[ $\ODEECGGenerator$ model training loss for MIT-BIH Arrhythmia dataset on PhysioNet First Channel]{\includegraphics[width=0.45\textwidth]{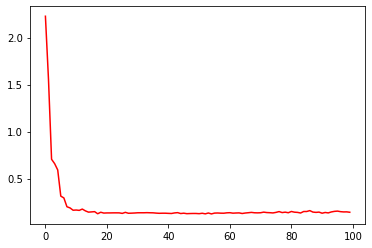}}\label{fig:loss_arr_f}
\hfill
\subfigure[ $\ODEECGGenerator$ model training loss for MIT-BIH Arrhythmia dataset on PhysioNet Second Channel]{\includegraphics[width=0.45\textwidth]{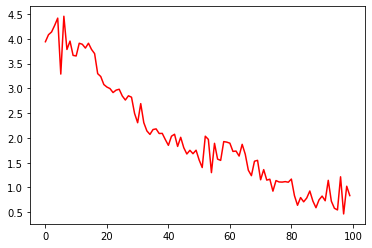}}\label{fig:loss_arr_s}
\vfill
\caption{Loss analysis for training  $\ODEECGGenerator$ model}
\label{fig:loss}
\end{figure*}

Fig.~\ref{fig:generated-signal}(b) and Fig.~\ref{fig:generated-signal}(c) shows the generated ECG for Normal and Arrhythmia signal by the proposed $\ODEECGGenerator$ model after training for 100 iterations.  Fig \ref{fig:loss} shows training loss for the generated ECG by by Model proposed in \cite{delaney2019synthesis} after training for 100 iterations.

The training loss analysis shown in \ref{fig:loss} shows that ODERNN based model reach a stable result within 20 iteration, which is significantly quicker than GAN based model \cite{delaney2019synthesis}.

\begin{figure*} [htb]
  \subfigure[Generator  training loss for MIT-BIH Normal Sinus Rhythm Database]{\includegraphics[width=0.5\textwidth]{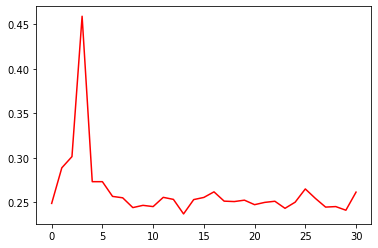}}\label{fig:gloss}
\hfill
\subfigure[Discriminator training loss for MIT-BIH Normal Sinus Rhythm Database]{\includegraphics[width=0.5\textwidth]{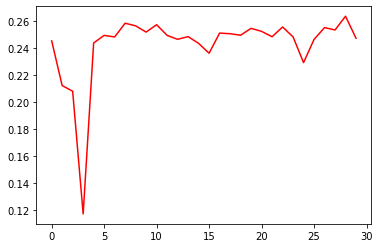}}\label{fig:dloss}
\vfill
\caption{Loss analysis for training GAN model}
\label{fig:gan_loss}
\end{figure*}

Similarly, Fig \ref{fig:gan_loss} shows the loss for Generator and Discriminator training for the GAN model over 100 epochs. 

The result evaluation between the proposed  $\ODEECGGenerator$ model and the existing GAN model shows that NODE enables  $\ODEECGGenerator$ to achieve comparatively higher accuracy with fewer parameters and a concise data series.   $\ODEECGGenerator$ model does not need to be trained for a longer time and essentially a hidden dimension. 

\begin{figure*} [htb]
  \subfigure[NeuralCDE generated ECG Signal]{\includegraphics[width=0.45\textwidth]{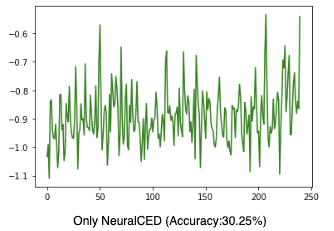}}\label{fig:neuralcde-perf}
\hfill
\subfigure[ODE GAN generated ECG Signal]{\includegraphics[width=0.45\textwidth]{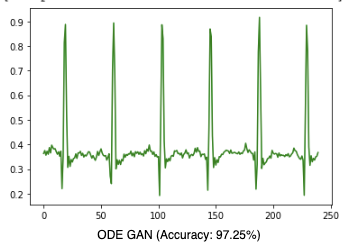}}\label{fig:odegan-perf}
\vfill
\subfigure[ODE GAN 2 generated ECG Signal]{\includegraphics[width=0.45\textwidth]{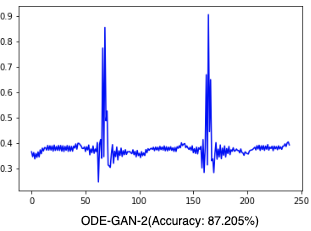}}\label{fig:odegan2-perf}
\caption{Generated ECG signals by different models}
\label{fig:generated-signal-2}
\end{figure*}

Fig. \ref{fig:generated-signal-2} shows the generated ECG by Neural CDE and proposed ODE GAN model with $\ODEECGGenerator$ as well as ODE GAN model with NODE based generator and discriminator. These generated ECG signals demonstrate that NODE based models can perform significantly better in continuous medical data generation.

\subsection{Comparative Analysis of GAN and proposed models for medical time series generation}

The normal sinus rhythm in the Normal Sinus Rhythm Database has a rate between 50 and 100 beats/minute at rest. This is the standard heart rate with a periodic pattern. There are two separate channels in the dataset. The first channel (Fig. \ref{norm-first}) contains ECG of continuous rhythm, where the Second Channel of the Normal Sinus Rhythm Database ((Fig. \ref{norm-second})) presents ECG with variant rhythm. The loss is shown in Fig. \ref{fig:loss_normal_f} and \ref{fig:loss_arr_f} shows that ECG with the minor variant in the series can quickly achieve high accuracy in the case of the  $\ODEECGGenerator$ model. When there is more variant in the time series, and the pattern changes frequently, it takes comparatively higher time to achieve acceptable accuracy.  
However, the training time for the  $\ODEECGGenerator$ model is still significantly lower than the GAN.

 $\ODEECGGenerator$ model achieve acceptable accuracy  for Arrhythmia dataset comparatively later than Normal Sinus dataset as shown in Fig. \ref{fig:loss_normal_f} and Fig. \ref{fig:loss_arr_f}.

A similar model, called ECG-ODE-GAN \cite{golany2021ecg} as proposed ODEGAN model described in \ref{sec:modelb} tries to learns purely data-driven dynamics from ECG signal. This  ECG-ODE-GAN model also shows the impact of physical parameters in morphological descriptors of the ECG signal instances.

For proposed NODE-based models, we observed better performance with batch size $>$ 50. If the batch size is too small, it creates unnecessary noise. The principle of the proposed NODE models is to learn the hidden dynamics of the continuous-time series from the pattern that represents the hidden dynamics most rather than considering each time steps $x_{t}$. As proposed NODE based models take the initial state $x_{0}$ of any batch as the input, it is redundant to use smaller batch sizes, e.g. 25, 20, 10. For the GAN model, we used a series of lengths 6446, but for the NODE based proposed model, we will use a series of 100 lengths. We observed that length series create instability during the training, and the model's performance drop significantly. GAN also requires a longer training time. Therefore, we avoid using length time series as training data. 
\clearpage
\section{Conclusion}

We have presented three different models for continuous medical data synthesis. Firstly, we introduce a new technique to learn the hidden dynamics of ECG signal by ODE-RNN only. Secondly, we introduced a generative adversarial network with a NODE Generator and a standard CNN based Discriminator. For the third model, we experimented with a NODE as Generator and NeuralCDE as the Discriminator. Finally,  we demonstrated that NODE models could improve the accuracy of standard generative adversarial networks. As NODE models are data-driven generative model, they perform better as the Generator in generative adversarial networks.  These hybrid Neural generative adversarial network can be used to generate any system that can de be described as ODE. The performance evaluation also shows that these models perform better when the batch size for training data is small (20-50) and fewer parameters. Furthermore, if the pattern in the signal is regular, the train time is lower than in the case of an irregular signal. This model can be a suitable candidate to generate systems described by PDEs as well, as they are data-driven and continuous in time.

\bibliographystyle{unsrtnat}
\bibliography{references}  

\end{document}